# Enabling Grasp Action: Generalized Evaluation of Grasp Stability via Contact Stiffness from Contact Mechanics Insight

Huixu Dong, Chen Qiu, Dilip K. Prasad, Ye Pan, Jiansheng Dai, *Fellow IEEE*, I-Ming Chen, *IEEE Fellow*

*Abstract*—**Performing a grasp is a pivotal capability for a robotic gripper. We propose a new evaluation approach of grasping stability via constructing a model of grasping stiffness based on the theory of contact mechanics. First, the mathematical models are built to explore "soft contact" and the general grasp stiffness between a finger and an object. Next, the grasping stiffness matrix is constructed to reflect the normal, tangential and torsion stiffness coefficients. Finally, we design two grasping cases to verify the proposed measurement criterion of grasping stability by comparing different grasping configurations. Specifically, a standard grasping index is used and compared with the minimum eigenvalue index of the constructed grasping stiffness we built. The comparison result reveals a similar tendency between them for measuring the grasping stability and thus, validates the proposed approach.**

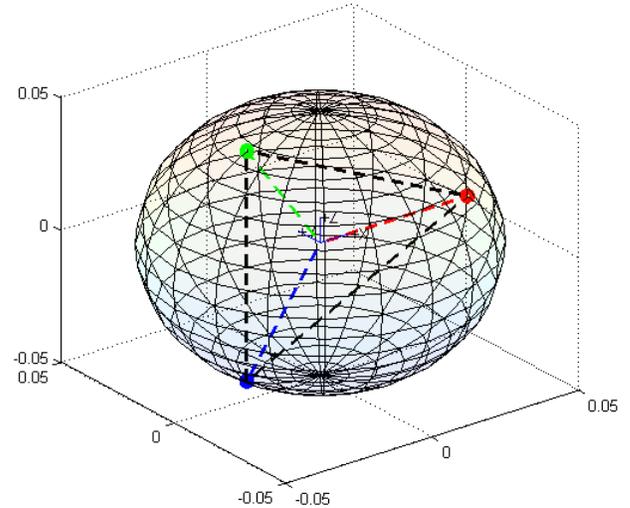

Fig. 1. Robotic gripper grasping a spherical object by three fingers.

## I. INTRODUCTION

THE grasp is a crucial capability for a robotic gripper[1]. Stability analysis is one of the foundational problems for robotic grasp. The formulation and characteristics of the stability measure thus play a key role in grasp tasks such as planning and executing a grasp, designing robotic hand[2-4]. It is common for soft-finger contact in grasping applications since soft-finger contact reflects a practical situation where a robotic finger contacts an object. Regardless of whatever grasp types, the prerequisite of a stable grasp is that the grasped object can keep a stable state of quasi-static equilibrium under a certain external disturbance.

Two main versatile approaches to measuring the grasping stability are as follows. The first is that the potential energy is applied to evaluating the grasping stability. The elastic system

Huixu Dong is with Robotics Institute, Carnegie Mellon University, PA 15213 USA (e-mail: huixud@andrew.cmu.edu).

Chen Qiu is with Centre for Robotics Research, King's University London, United Kingdom.

Dilip K. Prasad is with Nanyang Technological University, 639798 Singapore.

Ye Pan is with University College London, United Kingdom.

Jiansheng Dai is with Centre for Robotics Research, King's University London, United Kingdom.

I-Ming Chen is with Nanyang Technological University, 639798 Singapore (e-mail: michen@ ntu.edu.sg).

is used in optimizing the grasping quasi-static equilibrium models for analyzing the grasping stability[5]. The contact geometry that generates the important effect on grasping stability was investigated in [7, 8] based on the stiffness matrix. The generalization eigen-decomposition was used as a measure of grasp stability for arbitrary perturbations and loading conditions [9-11]. By building a generalized contact stiffness matrix, the authors explored the contact characteristics with line springs as the equivalence of the soft-finger contact, which reveals the rotational effects on the contact stiffness based on the screw theory for evaluating grasping stability [12, 13]. The other is that the form and force closures are applied to exploring a grasp stability[14, 15]. The following references are far from complete but somewhat representative for evaluating a grasp stability. As for evaluating a grasping stability, the authors constructed a mobility theory to present the effect of curvatures of contact surface and object from grasping form closure insight [16, 17]. The polyhedral bounds including contact forces, normals, curvatures at contact surfaces were used for evaluating grasping stability based on the grasp force closure [18]. The contact forces performing a grasp was decomposed via a coordinate system to explore the grasp stability in [19]. For [20],Tsuji et al. used a few ellipsoids to approximate the friction cone such as to test the force closure easily for a grasp.

To the best of our knowledge, few available published works take into consideration about the effects of contact mechanics on the grasp stability. As the main contribution of this work, we propose a new generalized and quantitative analysis of the grasp stability through providing an outline of measuring the grasp stability by means of the constructed grasp stiffness matrix. Here for each grasp contact, patch contact model instead of point contact model is adopted, thus normal force, tangential force and torsional moment are considered. First, we introduce an equivalent model of grasp contact and a general procedure of constructing grasping stiffness matrix, including extracting contact location and orientation information from six-dimensional force/torque sensor; constructing global grasping stiffness using contact stiffness coefficients and adjoint transformation matrix. Then, we explore the deduction of contact stiffness coefficients following contact mechanics modelling principals. Contact stiffness coefficients include normal stiffness coefficient, tangential stiffness coefficient and torsional stiffness coefficient, and they are modelled as functions of local material curvature, contact material properties as well as related force/torque magnitudes. Next, we construct the grasp stiffness matrix based on the models built above, and evaluate grasp stability using the minimum eigenvalue of constructed stiffness matrix. Finally, we design two grasp cases to verify the proposed criterion of grasping stability by comparing different grasping configurations. Specifically, a standard grasping index is used and compared with the eigenvalue index of the constructed grasping stiffness we built. The comparison result reveals a similar tendency between them and thus validate the proposed approach.

The rest of contents consist of five sections. Section II constructs an equivalent grasp model and general model of grasping stiffness. The construction of the contact stiffness and determination of the normal, tangential and torsion stiffness coefficients are illustrated in section III. The effects of factors on stiffness coefficients are discussed in detail in section IV. Section V describes the evaluations of grasp stability based on the constructed grasp stiffness, followed by conclusions and future work in Section VI.

## II. GRASP STIFFNESS CONSTRUCTION

### A. Problem Formulation

As stated above, the soft finger contact always occurs in life. That is, this soft contact is the most general case in practical grasps. For instance, humans grasp objects using this type frequently. A soft finger contact between two real (nonrigid, possibly inelastic) objects results in mutually transmitting a distribution of contact tractions that are compressive over a finite area of contact.

The instinct sensing is implemented to reflect the contact situations when a soft finger contact occurs. Specifically, a surface of the end of robotic finger, which we name it a fingertip, is attached by a six-dimension force/torque sensor. Referred to the reference frame $O_1$, the force/torque sensor can obtain all three components of both the resultant force $f$ and the resultant moment $m$ (see Fig. 2). Note that the choice of the reference frame $O_1$ is arbitrary, as we can easily express $f$ and $m$ in terms of any other coordinate frame fixed to $O_1$.

The original force sensor data (the wrench) from a fingertip is described as

$$w = [f \quad m]^T. \tag{1}$$

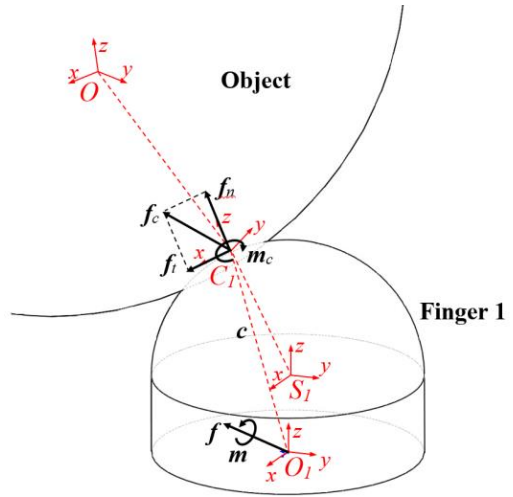

Fig.2. Force equilibrium of contact model. $\{O_1 - xyz\}$, $\{S_1 - xyz\}$, $\{C - xyz\}$ and $\{O - xyz\}$ denote the sensor frame, the fingertip frame, the contact frame and the object frame, respectively; $f_c$ and $m_c$ are the contact force and moment, respectively; $f_n$ and $f_t$ represent the normal and tangential forces in $\{C - xyz\}$. $c$ is the position vector between $\{O_1 - xyz\}$ and $\{C - xyz\}$.

The fingertip surface can be described by the implicit relation

$$S(i) = 0 \tag{2}$$

where $i$ is a point in space defined with respect to $O_1$. The surface $S$ should have continuous first derivatives, so that a normal unit vector $n$ can be defined at every point on $S$ as

$$n = \frac{\nabla S(i)}{\|\nabla S(i)\|} \tag{3}$$

from which $\nabla$ indicates the gradient operator. Let $C$ be the contact centroid. $f_c$ and $m_c$ represent the force and moment applied at $C$ respectively, which are equivalent to a "soft finger" contact. The measurable quantities $f$ and $m$ are related to the unknowns $c$, $f_c$ and $m_c$, by force and moment balance equations with respect to the coordinate frame of force-torque sensor,

$$\begin{cases} f = f_c; \\ m = m_c + c \times f_c. \end{cases} \tag{4}$$

When soft finger contacts exists in grasping, the torque $m_c$ and the unit vector $n$ are parallel being normal to the surface that haws the contact centroid; thus

$$n \propto m_c = \frac{K}{2}\nabla S(c) \tag{5}$$

for some constant $K$.

From which we are able to obtain the normal direction $n$ of contact area, as well as the location $C$ in the sensor coordinate frame $\{O_1 - xyz\}$. According to the vector projection, the normal force and tangent force can be developed as

$$\begin{cases} f_n = \frac{n(c)^T f}{n(c)^T n(c)} \cdot n(c), \\ f_t = f - f_n \end{cases} \tag{6}$$

To simplify derivations based on a closed-form algorithm, a specific class of surfaces-namely is applied to restricting the fingertip surfaces and thus, quadratic forms of the type

$$S(r) = i^T A^T A i - R^2 = 0 \tag{7}$$

where $A$ is a constant coefficient matrix, and $R$ is a scale factor used for convenience. Because the reference frame $O_1$ can be moved arbitrarily, we can assume without loss of generality that $A$ can be written in diagonal form

$$A = \begin{pmatrix} \frac{1}{\alpha} & 0 & 0 \\ 0 & \frac{1}{\beta} & 0 \\ 0 & 0 & \frac{1}{\gamma} \end{pmatrix} \quad (8)$$

with $0 < \frac{1}{\alpha} \leq 1, 0 < \frac{1}{\beta} \leq 1$ and $0 < \frac{1}{\gamma} \leq 1$. In this case, the principle axes of the ellipsoid form by the surface are given by $2R\alpha$, $2R\beta$ and $2R\gamma$, respectively.

Indeed, many researchers can use the intrinsic contact sensing to explore more complex contact surfaces rather than one with the simple geometry introduced above. However, the described model of sensing force/torque information at the surface contacts is suitable for compound convex surfaces consisting of simpler surfaces that share the same normal at the corresponding boundaries.

### B. Building Adjoint Transformation

Since a screw can be represented in the form of a six-dimensional vector, it follows certain rules of coordinate transformation when its based coordinate frame changes. Two Cartesian coordinate frames $\{A - x_a y_a z_a\}$ and $\{B - x_b y_b z_b\}$ are used to demonstrate the coordinate transformation of a screw, as shown in Fig. 3. Assume the symbol of a screw is $S_a$ in the coordinate frame $\{A - x_a y_a z_a\}$. Similarly, assume $S_b$ is the symbol of $S$ in the coordinate frame $\{B - x_b y_b z_b\}$ and it can be written as where both $S_a$ and $S_b$ are written using Plucker ray coordinates [21]. We can obtain the relationship between $S_a$ and $S_b$ as

$$S_a = Ad_{ab} S_b \quad (9)$$

where $Ad_{ab}$ is the adjoint transformation matrix and it has the form

$$Ad_{ab} = \begin{bmatrix} R_{ab} & 0 \\ P_{ab} R_{ab} & R_{ab} \end{bmatrix} \quad (10)$$

where $R_{ab}$ is the $3 \times 3$ rotation matrix from $\{A - x_a y_a z_a\}$ to $\{B - x_b y_b z_b\}$, $P_{ab}$ is the anti-symmetric matrix of translation vector $p_{ab}$, it can be written as the following,

$$P_{ab} = \begin{bmatrix} 0 & -p_z & p_y \\ p_z & 0 & -p_x \\ -p_y & p_x & 0 \end{bmatrix}, \quad (11)$$

from which $p_{ab} = \begin{bmatrix} p_x & p_y & p_z \end{bmatrix}^T$. As a result, the equations above give us the general form of screw coordinate transformation using the adjoint matrix $Ad_{ab}$.

Thus, if we define the global coordinate frame $\{O - xyz\}$ (generally located at the centre of grasped object or somewhere else), then based on the screw theory, from the global coordinate frame $\{O - xyz\}$ to the sensing coordinate frame $\{O_1 - xyz\}$ and from the sensing coordinate frame $\{O_1 - xyz\}$ to the contact coordinate frame $\{C - xyz\}$, we can get the adjoint transformation matrices respectively, as follows,

$$Ad_{oo_1} = \begin{bmatrix} R_{oo_1} & 0 \\ P_{oo_1} R_{oo_1} & R_{oo_1} \end{bmatrix},$$
$$Ad_{o_1 c} = \begin{bmatrix} R_{o_1 c} & 0 \\ P_{o_1 c} R_{o_1 c} & R_{o_1 c} \end{bmatrix}, \quad (12)$$

where $R_{oo_1}$ and $R_{o_1 c}$ are the 3 by 3 rotation matrixes representing coordinate frame $\{O_1 - xyz\}$ and $\{C - xyz\}$ with respect to $\{O - xyz\}$ and $\{O_1 - xyz\}$, respectively. $P_{oo_1}$ and $P_{o_1 c}$ are the anti-symmetric matrixes representing the cross products of position vectors $p_{oo_1}$ and $p_{o_1 c}$, respectively. Now the problem turns out to be finding the adjoint matrix $Ad_{o_1 c}$ between contact coordinate frame $\{C - xyz\}$ and $\{O_1 - xyz\}$. For the position vector $p_{o_1 c}$, we have $p_{o_1 c} = c$. Here we select Z-Y-Z Euler angle representation to derive $R_{o_1 c}$, which can be established as

$$R_{o_1 c} = R_z(\emptyset) R_y(\psi) R_z(\gamma) =$$
$$\begin{bmatrix} c\emptyset & -s\emptyset & 0 \\ s\emptyset & c\emptyset & 0 \\ 0 & 0 & 1 \end{bmatrix} \begin{bmatrix} c\emptyset & 0 & s\emptyset \\ 0 & 1 & 0 \\ -s\emptyset & 0 & c\emptyset \end{bmatrix} \begin{bmatrix} c\gamma & -s\gamma & 0 \\ s\gamma & c\gamma & 0 \\ 0 & 0 & 1 \end{bmatrix} \quad (13)$$

where the coordinate frame $\{C - xyz\}$ is defined such that the x-axis coincides with $f_t$, and z-axis coincides with $f_n$. Thus in $\{C - xyz\}$, the wrench $w_c$ is represented as

$$w_c = [f_c \quad m_c]^T = [f_t \quad 0 \quad f_n \quad 0 \quad 0 \quad m_c]^T \quad (14)$$

where $f_c$ and $m_c$ represent the loading force and moment in the contact area, respectively. And the force parts of $f_c$ and $f$ has the relationship

$$f = R_{o_1 c} f_c. \quad (15)$$

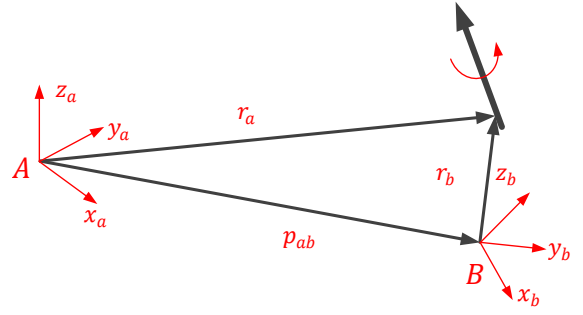

Fig. 3. Force equilibrium of contact model. $\{O_1 - xyz\}$, $\{S_1 - xyz\}$, $\{C - xyz\}$ and $\{O - xyz\}$ denote the sensor frame, the fingertip frame, the contact frame and the object frame, respectively; $f_c$ and $m_c$ are the contact force and moment, respectively; $f_n$ and $f_t$ represent the normal and tangential forces in $\{C - xyz\}$. $c$ is the position vector between $\{O_1 - xyz\}$ and $\{C - xyz\}$.

The first two rotation angles $\emptyset$ and $\psi$ can be obtained according to $n$ (which is already normalized) as

$$R_z(\emptyset) R_y(\psi) \begin{bmatrix} 0 \\ 0 \\ 1 \end{bmatrix} = n = \begin{bmatrix} n_x \\ n_y \\ n_z \end{bmatrix}. \quad (16)$$

$n$ is the normal vector described in $\{O_1 - xyz\}$ representing the z-axis direction of $\{C - xyz\}$, so we have

$$n = R_{o_1 c} \begin{bmatrix} 0 \\ 0 \\ 1 \end{bmatrix}$$
$$= R_z(\emptyset) R_y(\psi) R_z(\gamma) \begin{bmatrix} 0 \\ 0 \\ 1 \end{bmatrix} = R_z(\emptyset) R_y(\psi) \begin{bmatrix} 0 \\ 0 \\ 1 \end{bmatrix}) \quad (17)$$

And the third rotation angle can be obtained from the third equation, their analytical forms are as follows,

$$\psi = \cos^{-1}(n_z), \emptyset = \tan^{-1}\left(\frac{n_y}{n_x}\right),$$
$$\gamma = \cos^{-1}\left(-\left(\frac{f_t}{\|f_t\|}\right)_{(z)} \frac{1}{\sqrt{1 - n_z^2}}\right). \quad (18)$$

Thus, we are able to get the adjoint matrix $Ad_{o_1c}$ as well.

### C. Construction of Stiffness Matrix

We make an assumption that the contact normals point inward and consider the stiffness of the robotic fingertip as the equivalence of passive compression line springs, as shown in Fig. 4. Referring to [22, 23], we develop the contact stiffness matrix $K_c$ in the coordinate frame $\{C - xyz\}$, then integrate it into the global coordinate frame $\{O - xyz\}$ according to the equilibrium

$$K = \sum_{i=1}^{n} (Ad_{oo_i} Ad_{o_i s_i}) K_c (Ad_{oo_i} Ad_{o_i s_i})^T. \quad (19)$$

Since normal force, tangent force and normal torque is considered in this stiffness matrix (bending in $x$-axis and $y$-axis are ignored in coordinate frame $\{C - xyz\}$), the stiffness matrix $K_c$ should has the form

$$K_c = J \, diag([k_n, k_t, k_\tau]) J^T \quad (20)$$

where

$$J = \begin{bmatrix} 0 & 0 & 1 & 0 & 0 & 0 \\ 1 & 0 & 0 & 0 & 0 & 0 \\ 0 & 0 & 0 & 0 & 0 & 1 \end{bmatrix}^T.$$

We need to determine the stiffness coefficients $k_n$, $k_t$ and $k_\tau$. There are several approaches to finding the possible solutions, including analytical models, FEA simulations and Experiment tests. The appendix introduces the detail derivations.

## III. CONTACT STIFFNESS MODELLING

### A. Constructing Model of Elastic Half Space

The normal and tangential forces are applied to generating the stresses and deformations in a closed area $S$ of the surface in the neighbourhood of the origin for an elastic half-space bounded by the plane surface $z = 0$ [24], as shown in Fig. 5.

We denote by $C(\xi, \eta)$ a surface point in S, whilst $A(x, y, z)$ represents a point within the body of the solid. The distance between $C(\xi, \eta)$ and $A(x, y, z)$ is provided as

$$CA \equiv \rho = \{(\xi - x)^2 + (\eta - y)^2 + z^2\}^{1/2}. \quad (21)$$

The potential functions, each satisfying Laplace's equation, are defined as follows,

$$F_1 = \int_S \int q_x(\xi, \eta) \Omega d\xi d\eta$$

$$G_1 = \int_S \int q_y(\xi, \eta) \Omega d\xi d\eta$$

$$H_1 = \int_S \int p(\xi, \eta) \Omega d\xi d\eta \quad (22)$$

where $\Omega = z \ln(\rho + z) - \rho$, the normal force, x-axis tangential force and y-axis tangential force distributions are represented by $p(\xi, \eta)$, $q_x(\xi, \eta)$ and $q_y(\xi, \eta)$ acting on S, respectively.

$$F = \frac{\partial F_1}{\partial z} = \int_S \int q_x(\xi, \eta) \ln(\rho + z) \, d\xi d\eta$$

$$G = \frac{\partial G_1}{\partial z} = \int_S \int q_y(\xi, \eta) \ln(\rho + z) \, d\xi d\eta$$

$$H = \frac{\partial H_1}{\partial z} = \int_S \int p(\xi, \eta) \ln(\rho + z) \, d\xi d\eta. \quad (23)$$

We have

$$\psi_1 = \frac{\partial F}{\partial x} + \frac{\partial G}{\partial y} + \frac{\partial H}{\partial z} \quad (24)$$

and

$$\psi = \frac{\partial \psi_1}{\partial z} = \frac{\partial F}{\partial x} + \frac{\partial G}{\partial y} + \frac{\partial H}{\partial z} \quad (25)$$

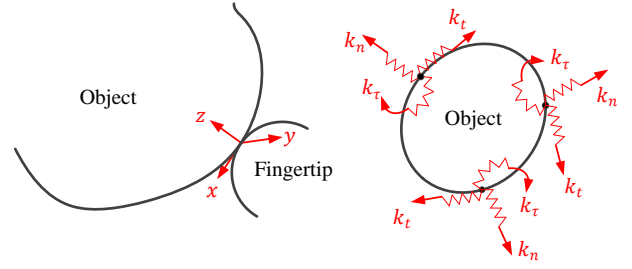

Fig. 4. The stiffness of the fingertip represented by a set of passive compression line springs.

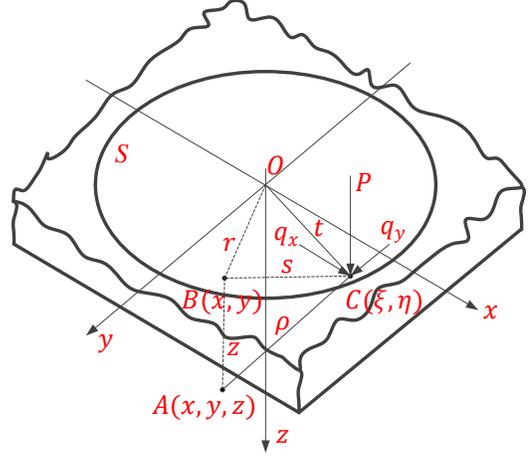

Fig. 5. The model of the elastic half-space.

The elastic displacements $u_x$, $u_y$ and $u_z$ at any point $A(x, y, z)$ in the solid body are expressed in terms of the above functions as follows:

$$u_x = \frac{1}{4\pi G} \left\{ 2 \frac{\partial F}{\partial z} - \frac{\partial H}{\partial x} + 2\nu \frac{\partial \psi_1}{\partial x} - z \frac{\partial \psi}{\partial x} \right\}$$

$$u_y = \frac{1}{4\pi G} \left\{ 2 \frac{\partial G}{\partial z} - \frac{\partial H}{\partial y} + 2\nu \frac{\partial \psi_1}{\partial y} - z \frac{\partial \psi}{\partial y} \right\}$$

$$u_z = \frac{1}{4\pi G} \left\{ \frac{\partial H}{\partial z} + (1 - 2\nu)\psi - z \frac{\partial \psi}{\partial z} \right\} \quad (26)$$

where $G$ is the shear modulus and $\nu$ represents the Poisson's ratio.

### B. Normal Force of Hertz Equations for Circular Contact

When constructing mathematical models in terms of evaluating grasping stability, we have to provide the corresponding assumptions. One is that the magnitude of contact area between two elastic solids is quite small compared to the dimension of objects and the radii of curvature. The other is that normal circular contact formed locally with orthogonal radii of curvature leads to a circular contact for simplifying the calculations when two elastic solid objects come into contact.

As shown in Fig. 6, at the beginning, without the load, two bodies just contact at one point(A). There are two points $M_1$ and $M_2$ being $r$ away from the common normal and being $z_1$, $z_2$ away from the tangential plane between two bodies, respectively. According to the geometric constraints, we can obtain the following equations,

$$\begin{cases} (R_1 - z_1)^2 + r^2 = R_1^2 \\ (R_2 - z_2)^2 + r^2 = R_2^2 \end{cases}, \quad (27)$$

If the points $M_1$ and $M_2$ is close, we have $z_1 \ll R_1$, $z_2 \ll R_2$, and thus,

$$\begin{cases} z_1 = \frac{r^2}{2R_1} \\ z_2 = \frac{r^2}{2R_2} \end{cases} \tag{28}$$

The distance between $M_1$ and $M_2$ is

$$z_1 + z_1 = \frac{r^2}{2R_c} \tag{29}$$

With $\frac{1}{R_c} = \frac{1}{R_1} + \frac{1}{R_2}$. Where $R_c$ represents the relative radius that expresses a summation of curvatures (or inverse radii). When the surface is convex, its curvature is positive while the curvature of concave surface is negative. Regardless of either the positive or negative symbols of radius, it represents an equivalent sphere in contact with a plane as long as $R_c$ is positive. As depicted in Fig.6(B), when a force $P$ is applied to loading along the normal, the local deformation results in a circular contact surface with the radius $a$ around the contact point. We denote by $\omega_1$ and $\omega_2$ the displacements along the $z_1$-axis and $z_2$-axis directions, respectively. The approximate distance $\delta$ between $O_1$ and $O_2$ is

$$\delta = z_1 + z_2 + \omega_1 + \omega_2. \tag{30}$$

According to the theory of elastic half space [25], the displacement of the point $M$, as shown in Fig.6(B), is under the normal force $q$ as follows,

$$\omega_1 = \frac{1-\nu_1^2}{\pi E_1} \int\int q \, ds d\psi \tag{31}$$

where $E_1, E_2$ are the elastic moduli; $\nu_1, \nu_1$ denote the Poisson's ratios associated with each body respectively. However, the integral should include the whole contact surface, similarly, the other displacement is described above. Thus,

$$\omega_1 + \omega_2 = \frac{1}{\pi E_c} \int\int q \, ds d\psi = \delta - \frac{r^2}{2R_c} \tag{32}$$

with $\frac{1}{E_c} = \frac{1-\nu_1^2}{E_1} + \frac{1-\nu_2^2}{E_2}$, where $E_c$ represents the contact modulus. We first have to calculate the normal force distribution $q$ for obtaining $\delta$. Based on the assumption of Hertz [26], the height of each point that rests on the half-sphere surface made along the boundary of contact surface represents the magnitude $h$ of the normal force $q$. Thus, the pressure force $q_o$ of the centre $O$ of the contact circular can be described as $q_o = ka$ where $k$ denotes the scale of the normal force distribution, as shown in Fig. 7. The normal force of a point in the contact circular is equal to the product of the height $h$ and the scale $k$ and thus,

$$\int q ds = \frac{q_o}{a} \int h ds = \frac{q_o}{a} A \tag{33}$$

with $A = \frac{\pi}{2}(a^2 - r^2 sin^2 \psi)$

where $A$ denotes the area of the half circular along the chord $mn$. Substituting $A$ into Eq., we can obtain the following equation,

$$\frac{1}{\pi E_c} \cdot 2 \cdot \int_0^{\frac{\pi}{2}} \frac{q_o}{a} \cdot \frac{\pi}{2}(a^2 - r^2 sin^2 \psi) d\psi = \delta - \frac{r^2}{2R_c} \tag{34}$$

Thus,

$$\frac{1}{\pi E_c} \cdot \frac{\pi q_o}{4a}(2a^2 - r^2) = \delta - \frac{r^2}{2R_c} \tag{35}$$

For the maximum normal force $q_o$, we integrate the total normal force $P$ within the half sphere as

$$q_o = \frac{3P}{2\pi a^2} \tag{36}$$

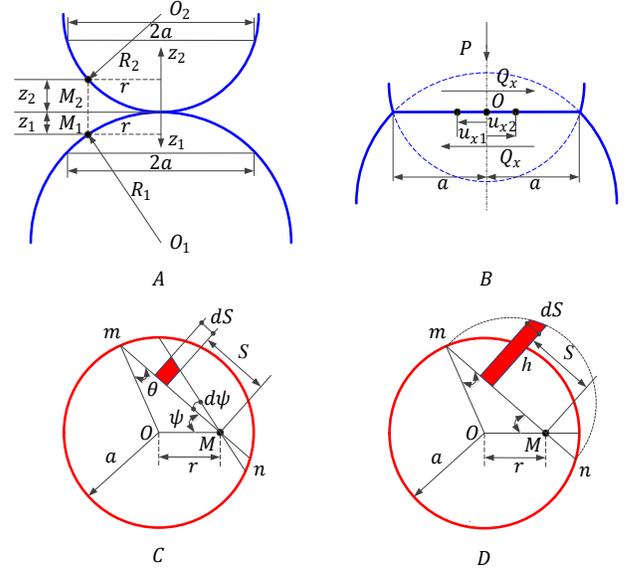

Fig. 6. Soft contact. At the beginning, without the load, two bodies just contact at one point(A). There are two points $M_1$ and $M_2$ being $r$ away from the common normal and being $z_1, z_2$ away from the tangential plane between two bodies, respectively. When a force $P$ is applied to loading along the normal, the local deformation results in a circular contact surface with the radius $a$ around the contact point(B). Due to the local deformation, $M_1$ and $M_2$ form the same point $M$ at the contact surface. $u_{x1}$ and $u_{x2}$ denote the displacements caused by the tangential force $Q_x$(C).

We obtain the radius $a$ of the circle that is related to the applied load $P$ by the equation,

$$a = \left(\frac{3PR_c}{4E_c}\right)^{\frac{1}{3}}. \tag{37}$$

The normal displacement $\delta$ is related to the maximum contact pressure by

$$\delta = \frac{a^2}{R_c} = \left(\frac{3P}{4E_c}\right)^{\frac{2}{3}} \left(\frac{1}{R_c}\right)^{\frac{1}{3}}. \tag{38}$$

Thus, we can obtain the normal stiffness as

$$k_n = \left(\frac{16PR_c E_c^2}{9}\right)^{\frac{1}{3}} \tag{39}$$

which expresses the elastic properties of both bodies effectively as a series combination of springs since stiffness is proportional to the elastic modulus for plain strain.

### C. Incipient Sliding of Elastic Bodies in Contact

A tangential force used for a stationary contact generates a relative tangential displacement governed principally by elastic deformation in the contact. Typically, small inelastic behaviour results from slip that always accompanies the elastic deformation. All hertz equations are applied along the normal direction for elastic contact. As the traction at the contact generates shear stress in the material, we can consider the contact shear modulus for simplifying the calculation. In the description the tangential traction has been assumed to have no effect upon the normal pressure.

A tangential force whose magnitude is less than the force of limiting friction ($Q < \mu P$, $\mu$ is the coefficient of friction), when applied to two bodies pressed into contact, will not give rise to a sliding motion, but nevertheless, will include frictional tractions at the contact interface.

Due to a tangential force $q_x(\xi,\eta)$ loading over the area $S$, the displacements and tangential stiffness are deduced. The tangential force $q_y$ along the $y$-axis and the normal pressure $p$ are both taken to be zero. Combining Eqs.(22-26) together, we can obtain

$$u_x = \frac{1}{4\pi G}\left\{2\frac{\partial^2 F_1}{\partial z^2} + 2\nu\frac{\partial^2 F_1}{\partial x^2} - z\frac{\partial^3 F_1}{\partial x^2 \partial z}\right\}. \quad (40)$$

When the appropriate derivatives are substituted in equations, we get

$$u_x = \frac{1}{4\pi G}\int_S \int q_x(\xi,\eta)\times M\,d\xi\,d\eta \quad (41)$$

With $M = \frac{1}{\rho} + \frac{1-2\nu}{\rho+z} + \frac{(\xi-x)^2}{\rho^3} - \frac{(1-2\nu)(\xi-x)^2}{\rho(\rho+z)^2}$.

If a tangential force $Q$ causes elastic deformation without slip at the interface, then the tangential displacement of any point in the contact area is the same. If $Q$ acts on the load area $S$ along the $x$-axis, this tangential displacement must also be parallel to the $x$-axis.

$$q_x(r) = q_0\left(1 - \frac{r^2}{a^2}\right)^{-\frac{1}{2}}, \quad (42)$$

with $q_0 = \frac{Q_x}{2\pi a^2}$ due to a concentrated tangential force $Q_x = q_x d\xi d\eta$ acting at $C(\xi,\eta)$.

Restricting the discussion to surface displacements within the loaded circle ($r \leq a$), Eq. (41) is reduced to

$$\bar{u}_x = \frac{1}{2\pi G}\int_S \int q_x(\xi,\eta)\left\{\frac{1-\nu}{s} + \frac{(\xi-x)^2}{s^3}\right\}d\xi\,d\eta \quad (42)$$

where $s^2 = (\xi-x)^2 + (\eta-y)^2$.

We transfer the coordinates from $(\xi,\eta)$ to $(s,\phi)$ to realize the surface integration as follows,

$$\xi^2 + \eta^2 = (x + s\cos\phi)^2 + (y + s\sin\phi)^2$$

$$q_x(s,\emptyset) = q_0 a(\alpha^2 - 2\beta s - s^2)^{-\frac{1}{2}} \quad (43)$$

with $\alpha^2 = a^2 - x^2 - y^2$ and $\beta = x\cos\phi + y\sin\phi$. Equation (42) then become

$$\bar{u}_x = \frac{1}{2\pi G}\int_0^{2\pi}\int_0^{s_1} q_x(s,\phi)\{(1-\nu) + \nu\cos^2\phi\}\,d\phi\,ds. \quad (44)$$

The limit $s_1$ is given by point D lying on the boundary of the circle, for which

$$s_1 = -\beta + (\alpha^2 + \beta^2)^{1/2}. \quad (45)$$

When performing the integration with respect to $\phi$ between the limits 0 and $2\pi$, so that for ($r \leq a$),

$$\bar{u}_x = \frac{q_0 a}{4G}\int_0^{2\pi}\{(1-\nu) + \nu\cos^2\phi\}\,d\phi = \frac{\pi(2-\nu)}{4G}q_0 a. \quad (46)$$

Under the action of the tangential force, the relative tangential displacement $\delta_x$ between two bodies is as follows,

$$\delta_x = u_{x1} - u_{x2} = \frac{Q_x}{8a}\left(\frac{2-\nu_1}{G_1} + \frac{2-\nu_2}{G_2}\right) \quad (47)$$

where $G_1$, $G_2$ represent the shear moduli and $\nu_1$, $\nu_2$ denote the Poisson's ratios of the two bodies, respectively.

The tangential displacement is directly proportional to the tangential force. This is unlike the normal approach of two elastic bodies which varies in a nonlinear way with normal load because the contact area grows as the load is increased. The tangential stiffness $k_t$ is provided depending on the Hooke's law $k_t = \frac{Q_x}{\delta_x}$ as follows,

$$k_t = 8a\left(\frac{2-\nu_1}{G_1} + \frac{2-\nu_2}{G_2}\right)^{-1}. \quad (48)$$

Any attempt to increase the tangential force $Q$ in excess of the friction force $\mu P$ causes the contact to slide.

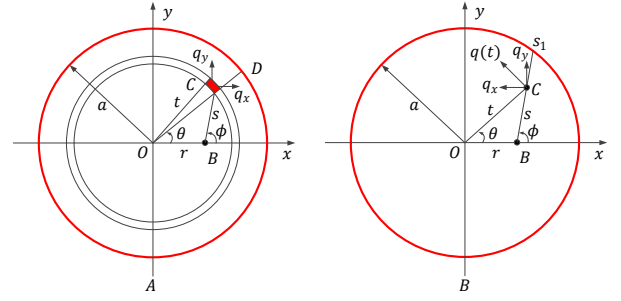

Fig. 7. Force equilibrium of contact model. $\{O_1 - xyz\}$, $\{S_1 - xyz\}$, $\{C - xyz\}$ and $\{O - xyz\}$ denote the sensor frame, the fingertip frame, the contact frame and the object frame, respectively; $f_c$ and $m_c$ are the contact force and moment, respectively; $f_n$ and $f_t$ represent the normal and tangential forces in $\{C - xyz\}$. $c$ is the position vector between $\{O_1 - xyz\}$ and $\{C - xyz\}$.

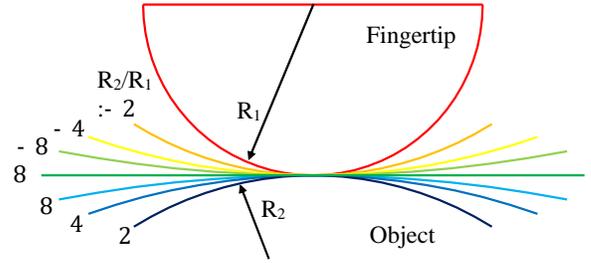

Fig. 8. Fingertip contact model with objects, the objects have different local curvatures.

### D. Torsion of Elastic Bodies in Contact

We investigate tangential forces acting on the load area $S$ in a circumferential direction which is perpendicular to the radius. A situation which is qualitatively similar to those discussed in the previous section occurs when two elastic bodies are pressed together by a normal force and then subjected to a varying twisting or "spinning" moment about the axis of their common normal. The twisting moment causes one body to rotate around the z-axis through a small angle $\beta$ relative to the other. Slip at the interface is resisted by frictional traction. Under the action of a purely twisting couple $M_z$ the state of each body is purely torsional.

For the circular region shown in Fig. 7(B) we shall assume that the magnitude of the traction $q(r)$ is a function of $r$ only. Thus

$$q_x = -q(r)\sin\theta = -q(t)\frac{\eta}{t};$$
$$q_y = q(r)\cos\theta = q(t)\frac{\xi}{t}. \quad (50)$$

The displacements $u_x, u_y$ and $u_z$ are described in the form of Eq.(26), where $H = 0$ and $F$, $G$ are given by

$$F = -\int_S \int \frac{q(t)}{t}\eta\ln(\rho+z)\,d\xi\,d\eta;$$
$$G = -\int_S \int \frac{q(t)}{t}\xi\ln(\rho+z)\,d\xi\,d\eta. \quad (51)$$

Due to the reciprocal nature of $F$ and $G$ related to coordinates, we have $\frac{\partial G}{\partial y} = -\frac{\partial F}{\partial x}$ and thus, the displacements on the surface can be simplified as follows,

TABLE I. Properties of fingertips

| | Material | Young's modulus $E$(pa) | Shear modulus $G$(pa) | Passion ratio | Contact radius(mm) |
|---|---|---|---|---|---|
| Fingertip | Rubber | 2.5e6 | 8.3e5 | 0.5 | 10 |
| Object | Rubber | 2.5e6 | 8.3e5 | 0.5 | $[-\infty, -20], [20, \infty]$ |
| | Polyethylene | 1.1e9 | 3.87e8 | 0.42 | $[-\infty, -20], [20, \infty]$ |
| | Aluminium | 7.1e10 | 2.67e10 | 0.33 | $[-\infty, -20], [20, \infty]$ |

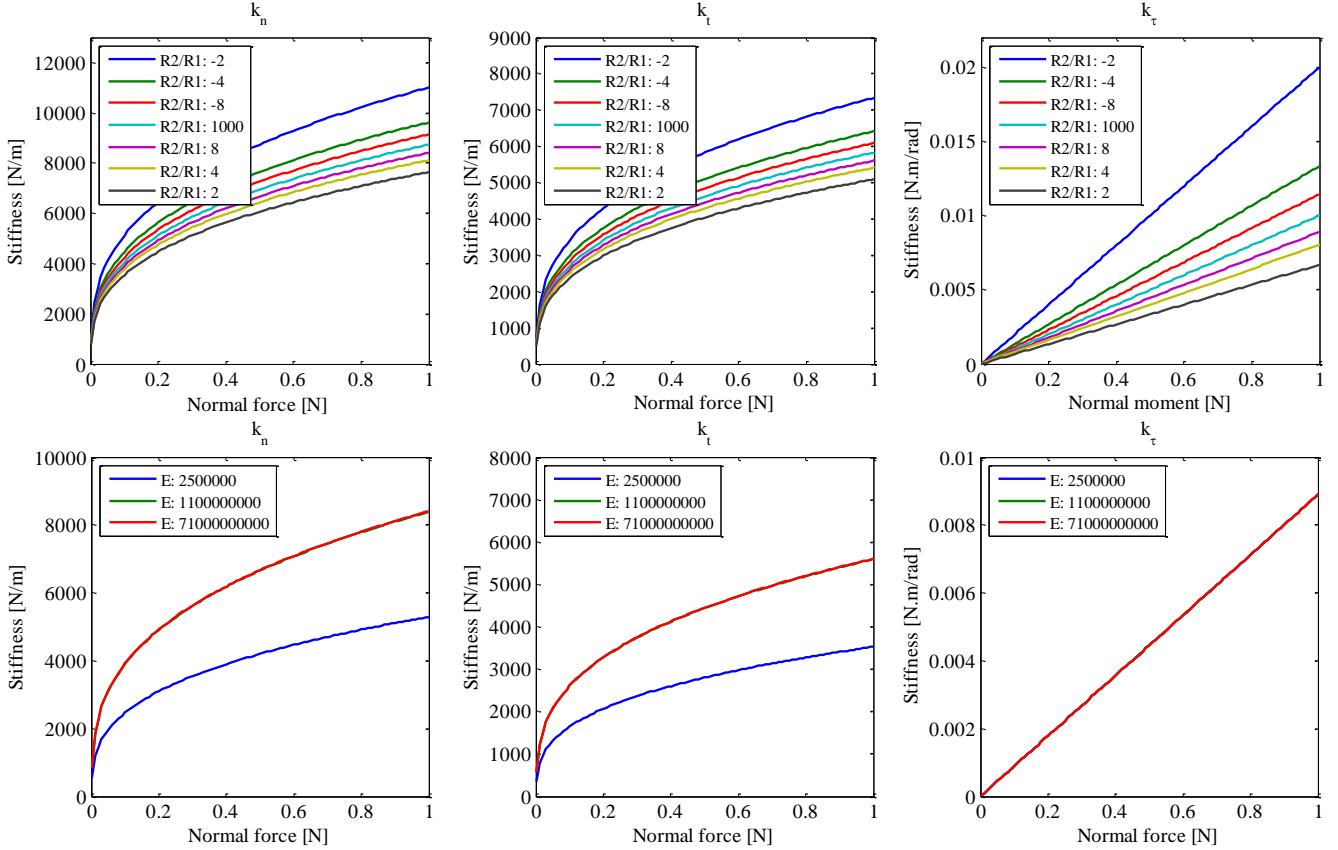

Fig. 9. Effects of object local curvatures and object materials on the stiffness coefficients $k_n, k_t, k_\tau$.

$$\bar{u}_x = \frac{1}{2\pi G}\frac{\partial F}{\partial z} = -\frac{1}{2\pi G}\int_0^{2\pi}\int_0^{S_1}\frac{q(t)}{t}\eta \, ds d\phi;$$
$$\bar{u}_y = \frac{1}{2\pi G}\frac{\partial G}{\partial z} = \frac{1}{2\pi G}\int_0^{2\pi}\int_0^{S_1}\frac{q(t)}{t}\xi \, ds d\phi;$$
$$\bar{u}_z = 0. \qquad (52)$$

As shown in Fig. 7(B), due to the point $B(x,0)$ and $\frac{\eta}{t} = \sin\phi$, the displacement $\bar{u}_x$ in Eq.(52) vanishes. Finally, we just have the component $\bar{u}_y$ expressed in a purely torsional deformation.

The force distribution to produce a rigid rotation of a circular region is provided as

$$q(r) = q_0 r(a^2 - r^2)^{-\frac{1}{2}}, \quad r \le a \qquad (53)$$

with $q_0 = \frac{3M_z r}{4\pi a^3}$, where $q(r)$ acts in a circumferential direction at all points in the contact circle.

Substituting in Eq.(52), we can obtain the surface displacement as

$$\bar{u}_y = \frac{q_0}{2\pi G}\int_0^{2\pi}\int_0^{S_1} N \, ds d\phi \qquad (54)$$

with $N = (a^2 - x^2 - 2xs\cos\phi - s^2)^{-\frac{1}{2}}(x + s\cos\emptyset)$.

The integral form can be given as

$$\bar{u}_y = \frac{\pi q_0 x}{4G} \qquad (55)$$

and thus, in view of the circular symmetry we can write

$$\bar{u}_\theta = \frac{\pi q_0 r}{4G}.$$

The force in Eq.(53) leads to a resultant twisting moment

$$M_z = \int_0^a q(r) 2\pi r dr = \frac{4}{3}\pi a^4 q_0. \qquad (57)$$

For one body, we have $\bar{u}_{\theta 1} = \beta_1 r$. Thus, the moment produces a rotation of the loaded circle through a resultant angle $\beta$ which is given by

$$\beta = \beta_1 + \beta_2 = \frac{3}{16}\left(\frac{1}{G_1} + \frac{1}{G_2}\right)\frac{M_z}{a^3} \qquad (58)$$

where $\beta_1$, $\beta_2$ represent the rotation angles of two bodies; $G_1, G_2$ denote the shear moduli of two bodies, respectively. Due to the Hooke's law $k_\tau = \frac{M_z}{\beta}$, the torsional stiffness is

$$k_\tau = \frac{16}{3}a^3\left(\frac{1}{G_1} + \frac{1}{G_2}\right)^{-1}. \qquad (59)$$

## IV. EFFECTS OF FACTORS ON STIFFNESS COEFFICIENTS

In this section, the stiffness coefficients developed in the above section are evaluated with respect to the material and geometrical properties of the contact surface of the gripper and grasped objects. The fingertip is assumed to be spherical using rubber materials, and its properties are listed in Table. I. In terms of the grasped object, they are selected to have various material properties such as rubber, polyethylene as well as aluminium, and different local contact curvatures.

We note that the differences between curvature and radius, curvature is the signed inverse of the radius of curvature at the point of contact, positive for convex surfaces. Fig.8 illustrates that the fingertip contacts objects with various local curvatures. The radius of fingertip and object are $R_1$ and $R_2$ respectively, and $R_2 > 0$ when it's convex and $R_2 < 0$ when it's concave. $R_2 = \infty$ when object is flat.

As shown in Fig.9, according to the models of the effects of the materials and local curvatures of the objects on the stiffness coefficients $k_n, k_t$ and $k_\tau$, these three stiffness coefficients improve with the normal force $P$ increasing for the fixed radius ratio and materials. Similarly, with the Young's modulus and shear modulus increasing, these stiffness coefficients will become larger. The contact between the fingertip and an object with a negative radius ratio results in higher stiffness coefficients compared to the contact with a positive radius ratio. The comparison between $k_n$, $k_t$ and $k_\tau$ indicates that the magnitudes of $k_n$ and $k_t$ are much larger than that of $k_\tau$ for the same materials and radius ratio. For $k_n$ and $k_t$, the values resting on small normal force ranges with less than 0.2N rise more rapidly than the values that fall into the big ranges with more than 0.2N. The variation of $k_n$ and $k_t$ is nonlinear as materials and local curvatures change, while $k_\tau$ varies at a linear mode with local curvatures and Young's modulus, shear modulus changing.

## V. EXPERIMENTS AND DISCUSSIONS OF GRASPING STIFFNESS EVALUATION

A cylindrical or spherical object is considered as a general represent object [4, 27] in the geometric models due to the following reasons. The surface of a cylindrical or spherical object is continuous and convex so that each link just has at most one contact point. Moreover, we can use a cylindrical or spherical object with just one variable (a radius) to simplify the geometrical model formulation and calculation.

We build a compliance model at each contact to measure the stability of the grasping system under small perturbations. The Cartesian stiffness matrix at each contact is applied to describing the force-displacement characteristics. When a grasp is regarded as a potential system, the matrix with second partial derivatives of the associated potential energy provides us insight into the grasping stability. If this matrix, also called the grasp stiffness matrix [4], is positive definite, the grasp is stable being subjected to small disturbance. It is required for a higher-order analysis by a positive semi-definite matrix. An indefinite matrix reveals that the grasp is unstable. Variants of this basic idea are used in definitions of first and second order stability [1, 4].

Here we use a classic example of three fingertip grasping to evaluate the effectiveness of the stiffness-matrix based stability evaluation approach. As shown in Fig.10, three fingertips are applied to grasping a spherical object which is a typical application scenario adopted in other researches[28]. The fingertip material is selected to be soft material that has the same material property used in [29]. The spherical object material is assumed to be aluminum without the loss of generality, as found in Table. I. In addition, the radius of fingertip is 10mm; the spherical object has various local contact curvatures, but the minimum distance between the contact point to the center of object remains the same, which is 40mm.

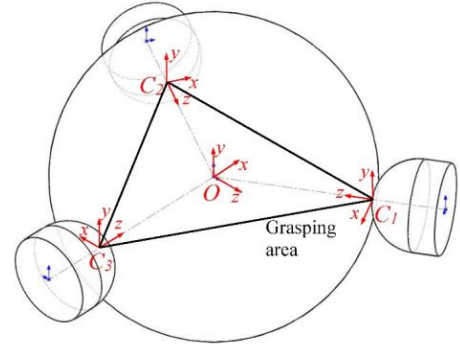

Fig. 10. Three finger grasping configuration.

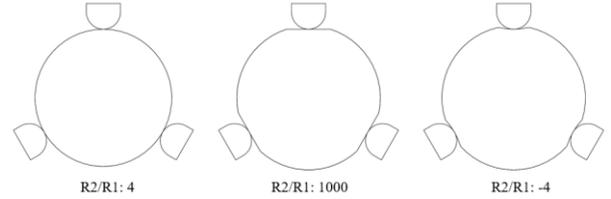

Fig. 11. Grasping configurations for different ratios of the fingertip radius and the object radius.

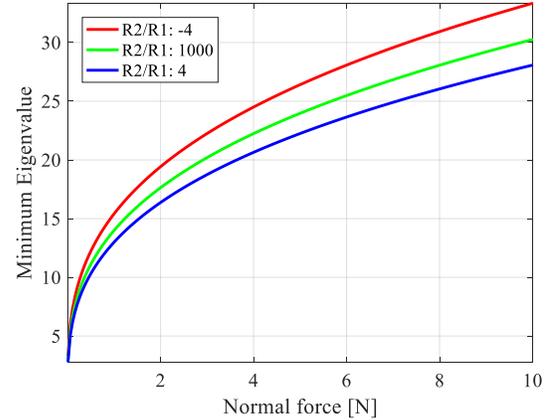

Fig. 12. Comparison of minimum eigenvalues with various contact local curvatures.

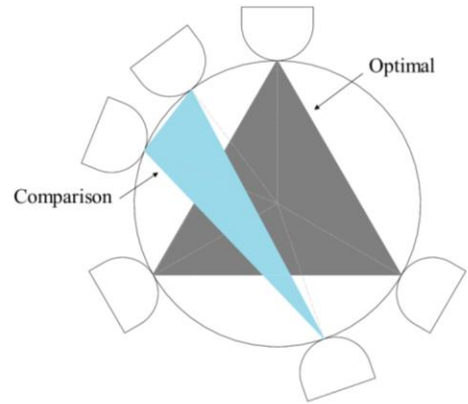

Fig. 13. Three finger grasping configuration.

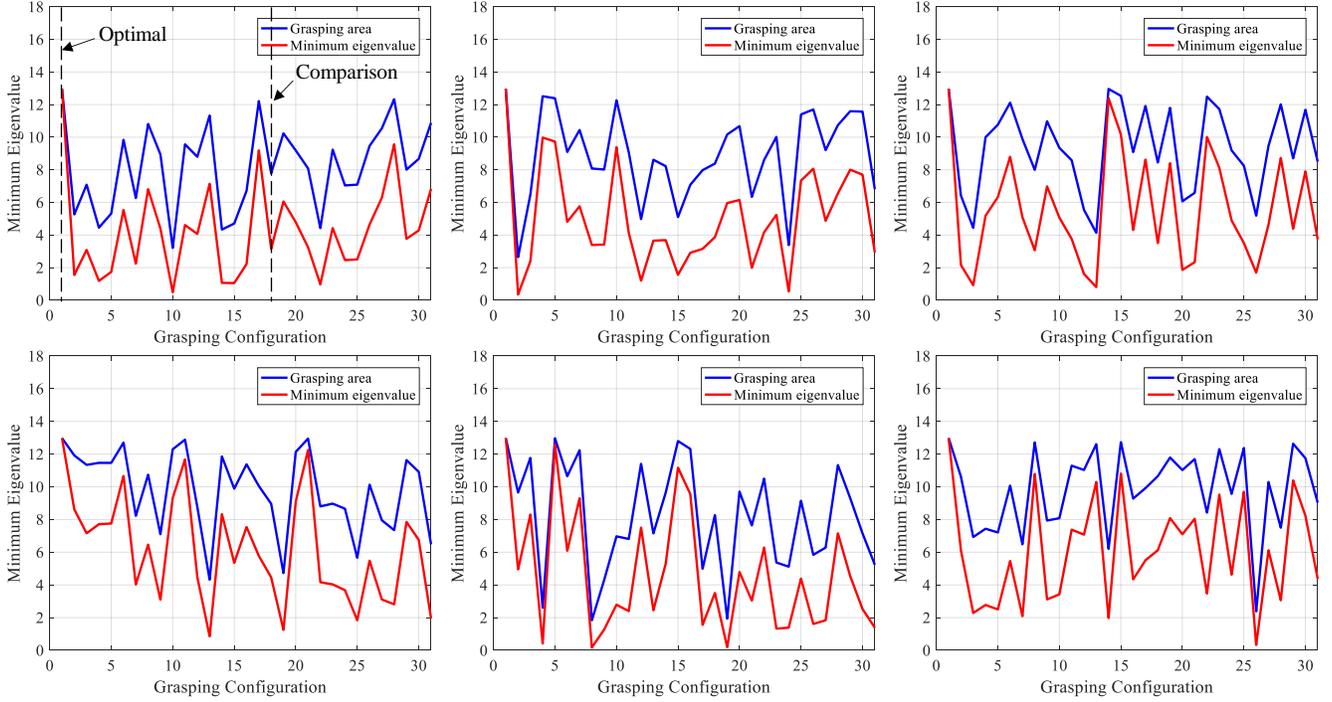

Fig. 14. Comparison of mini. eigenvalue with various contact local curvatures.

Referred to the derivation of the contact stiffness coefficients, the contact stiffness matrices are integrated into the global grasping stiffness matrix using adjoint coordinate transformation.

Global coordinate frame $\{O - xyz\}$ is attached at the centre of mass, and local contact coordinate frame $\{C_i - xyz\}(i = 1,2,3)$ is attached at each contact point with z-axis pointing to the centre of mass for a purpose of simplification(see Fig. 10). Thus the grasping stiffness matrix can be written as

$$K = \sum_{i=0}^{3} \left(Ad_{c_i o}\right)^{-1} K_c \left(Ad_{c_i o}\right)^{-T};$$

$$Ad_{c_i o} = \begin{bmatrix} R_{c_i o} & 0 \\ P_{c_i o} R_{c_i o} & R_{c_i o} \end{bmatrix}. \quad (60)$$

The positions of three fingertips can be obtained using active coordinate transformation from $\{C_i - xyz\}$ to $\{O - xyz\}$. The initial coordinate transformations of them are defined as

$$R_{c_i o} = R_y \left(\frac{2\pi}{3}(i-1) + \frac{\pi}{2}\right);$$
$$P_{c_i o} = [0 \quad 0 \quad R_2]^T \quad (61)$$

where $R_y$ and $R_2$ represent the general rotation and the displace from the centre of object and the contact position, respectively. For the construction of the grasping stiffness matrix $K$, it is noticed that its properties are determined by two factors, including the magnitude of normal force $f_n$ which determines the values of contact stiffness coefficients, as well as the spatial configuration which is represented by $Ad_{c_i o}$. In accordance with this, their effects are examined separately on the properties of $K$.

Further, following the criteria of evaluating grasping stability proposed in [28], we utilize the area enclosed by three contact points to be the benchmark function and compare with the minimum eigenvalue of stiffness matrix of each configuration built by ourselves. It is noticed that the optimal solution is the symmetric grasp with three contact points located on a big circle, thus we will pay special attention to this configuration and verify whether it also preserves the biggest index of our approach.

*Case A: Grasping stability based on minimum eigenvalue comparisons with various contact local curvatures*

The first case is completed by evaluating fingertip grasps with three different local contact curvatures at the optimal grasping configuration, which is shown in Fig. 11. The contact curvatures include normal outbound surface, flat surface and inbound surface. In addition, for each configuration, the contact force increases from 0N to 10N, and the minimum eigenvalue of constructed grasping stiffness matrix is obtained accordingly. The comparison result is further shown in Fig. 12. From the comparison result, we can see the minimum eigenvalue increases with the growth of the contact force. Then comparing the effects of local contact curvatures, we identify that the one with inbound surfaces results in the biggest mini. Eigenvalue to realize the best grasping stability among three different classic grasping configurations, which agrees with our intuition.

*Case B: Evaluation comparisons based on the minimum eigenvalue and the enclosed area proposed in [28]*

The second comparison is completed by comparing fingertip grasps with various grasping configurations using the standard spherical object. To evaluate the effectiveness for each grasping configuration, we compare the minimum eigenvalue of constructed grasping stiffness matrix with the grasping area [28]. Since the symmetric grasp with three contact points located on a big circle of the spherical object is identified as the optimal solution in terms of the grasping area, we would like to verify whether it leads to the biggest minimum eigenvalue index or not. Without the loss of generality, a total number of 31 types of grasping configurations in the big circle of the spherical object were selected, with their grasping areas and minimum eigenvalues compared. Fig. 13 presents two types of grasping, the one with grey area represents the optimal grasp while the comparison grasp with light blue area is selected from 1 of the other 30 grasping configurations which gives a relative small grasping area.

Further, the selected 31 types of grasping configurations are compared in Fig.14. The 31 grasping configurations are determined as follows. The $1^{st}$ configuration is the optimal grasping configuration, and the rest 30 configurations are generated using the rand algorithm. We make an additional modification by equaling the value of the grasping area and minimum eigenvalue of grasping stiffness matrix in the optimal configuration. We repeat 6 groups of comparison experiments by choosing different grasp configurations randomly. As illustrated in Fig.14, for all the comparison experiments, there are the consistencies between grasping area index and minimum eigenvalue index through all grasping configurations. In addition, both the grasping area and minimum eigenvalue of the symmetric grasping configuration can achieve the highest values among all the values, which verifies that such grasp can realize the best grasping stability.

## VI. CONCLUSION AND FUTURE WORK

A quantitative analysis of grasp stability is presented and discussed via constructing the grasp stiffness. The presented evaluation model is applicable to assess the stabilities of the fingertip grasp. The proposed approach of evaluating grasping stability is verified by comparing with the traditional method based on the grasping area. As to the future work, a mathematical model will be explored to describe the combined effects of the contact forces, the numbers of the contact points and the enveloping angle on a stable grasp.

## APPENDIX

For a robotic platform, a spatial force can be described using a wrench in screw theory. It contains a linear component (pure force) and an angular component (pure moment), which has the form as

$$w = \begin{bmatrix} f \\ m \end{bmatrix} \tag{A-1}$$

where $w$ is a $6 \times 1$ vector whose primary part $f = [f_x, f_y, f_z]^T$ is a $3 \times 1$ force vector and the second part $m = [m_x, m_y, m_z]^T$ is a $3 \times 1$ moment vector. The twist $T$ is provided as

$$T = \begin{bmatrix} \delta \\ \theta \end{bmatrix} \tag{A-2}$$

where $\theta = [\theta_x, \theta_y, \theta_z]^T$ is a $3 \times 1$ rotational displacement vector and $\delta = [\delta_x, \delta_y, \delta_z]^T$ is a $3 \times 1$ translational displacement vector. both $w_a$ and $w_b$ are written using the Plucker ray coordinates. The external force $w$ and the deformation twist $T$ are written using Plucker axis coordinates. $\Delta$ is the elliptical polar operator[30] as

$$\Delta = \begin{bmatrix} 0 & I_3 \\ I_3 & 0 \end{bmatrix} \tag{A-3}$$

$\Delta$ has some properties as follows,

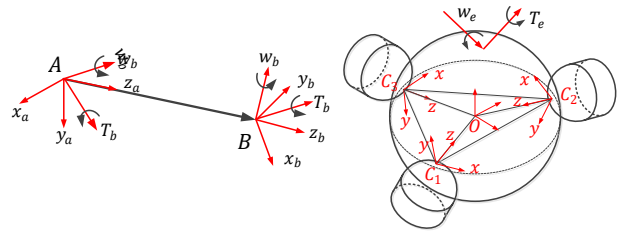

Fig. 15. Fingertip contact model with objects, the objects have different local curvatures.

$$\begin{cases} \Delta = \Delta^{-1} \\ \Delta = \Delta^T \\ \Delta\Delta = I_3 \end{cases} \tag{A-4}$$

For the adjoint matrix $Ad_{ab}$ stated above, it has the following properties as

$$\begin{aligned} Ad_{ab} &= \Delta(Ad_{ab}^{-T})\Delta \\ Ad_{ab} &= \Delta(Ad_{ab}^{-1})\Delta \\ Ad_{ab}^T \Delta Ad_{ab} &= \Delta \end{aligned} \tag{A-5}$$

### A. Construction of Global Stiffness Matrix

As shown in Fig.15, the external force $w$ and the resulted deformation twist $T$ are presented in the coordinate frame $\{B, x_b, x_b, x_b\}$, which are symbolized as $w_b$ and $T_b$, the relationship is provided as

$$T_b = C_b w_b \tag{A-6}$$

where $C_b$ is the compliance matrix in the coordinate frame $\{B, x_b, x_b, x_b\}$ [22]. The external load and deformation twist are written in the new coordinate frame as $w_a$ and $T_a$, as shown in Fig.15. The relationship between $w_a$ and $w_b$ can be written as

$$w_a = Ad_{ab}w_b \tag{A-7}$$

Also, for the deformation twist $T_a$ and $T_b$ we have a similar coordinate transformation formula as

$$\Delta T_a = Ad_{ab}\Delta T_b \tag{A-8}$$

Depending on the properties of $\Delta$ and $Ad_{ab}$ shown in Eq. (A-3) and Eq. (A-5), we simply Eq. (A-8) as

$$T_a = (\Delta Ad_{ab}\Delta)T_b = Ad_{ab}^{-T}T_b \tag{A-9}$$

Substituting Eq. (A-7) and Eq. (A-9) into Eq. (A-6), we can obtain

$$Ad_{ab}^T T_a = C_b Ad_{ab}^{-1} w_a \tag{A-10}$$

Further,

$$T_a = Ad_{ab}^{-T} C_b Ad_{ab}^{-1} w_a \tag{A-11}$$

Since the compliance matrix $C$ in the coordinate frame $\{A, x_a, x_a, x_a\}$ has the form $T_a = C_a w_a$, we can then get the relationship between $C_a$ and $C_b$ as

$$C_a = Ad_{ab}^{-T} C_b Ad_{ab}^{-1} \tag{A-12}$$

We simply by just reversing Eq. (A-12) as

$$C_a^{-1} = Ad_{ab} C_b^{-1} Ad_{ab}^T \tag{A-13}$$

According to the relationship between stiffness and compliance matrix $C = K^{-1}$, Eq. (A-12) can be further written as

$$K_a = Ad_{ab} K_b Ad_{ab}^T \tag{A-14}$$

### B. Construction of Global Grasping Stiffness Matrix

When an external load $w_e$ is applied at an object, a deformation $T_i(i = 1, ..., m)$ and a displacement $T_e$ of the object arises at the contact area. Thus, the relationship between $T_e$ and the elements $T_i(i = 1, ..., m)$ can be provided as

$$T_e = \sum_{i=0}^{m} T_i \tag{A-15}$$

which indicates $T_e$ is the aggregation of $T_i$ in the same global coordinate frame $\{O, x, y, z\}$. We can also use a global compliance matrix $C_e$ to establish the relationship between the twist $T_e$ and the wrench $w_e$ as

$$T_e = C_e w_e. \tag{A-16}$$

According to the coordinate transformation law, the deformation $T_i$ of the $i$-th flexible element can be represented in the local coordinate frame $\{O_i, x_i, y_i, z_i\}$ and it is symbolized as $T_i'$. Similar to Eq.(A-9), the relationship between $T_i$ and $T_i'$ can be written as

$$T_i' = Ad_{ie}^{-T} T_i;$$
$$Ad_{ie}^T T_i' = T_i. \tag{A-17}$$

Correspondingly, the relationship between $w_i$ and $w_i'$ is also provided as

$$w_i = Ad_{ie}^{-1} w_i' \tag{A-18}$$

where $Ad_{ie}(i = 1, \dots, m)$ is the adjoint transformation matrix between the local coordinate frame $\{O_i, x_i, y_i, z_i\}$ and the global coordinate frame $\{O, x, y, z\}$, it has the form by combining Eq.(A-15) and Eq.(A-17) as

$$T_e = \sum_{i=0}^{m} Ad_{ie}^T T_i'. \tag{A-19}$$

In contrast, the external load applied at the end effector is transmitting to each compliant element

$$w_i' = Ad_{ie} w_e \tag{A-20}$$

where $w_e$ is the external load in the global coordinate frame, $w_i'$ is the transmitted internal load applied at the $i$-th flexible element which is expressed in the local coordinate frame $\{O_i, x_i, y_i, z_i\}$. We have the relationship between $w_i'$ and $T_i'$ through the compliance matrix as

$$T_i' = C_i w_i' \tag{A-21}$$

where $C_i$ is the compliance matrix. Substituting Eq. (A-21) and Eq. (A-16) into Eq. (A-19), we can obtain

$$C_e T_e = \sum_{i=0}^{m} Ad_{ie}^T C_i w_i' \tag{A-22}$$

which can be further deduced by substituting into Eq. (A-20) as,

$$C_e w_e = \sum_{i=0}^{m} Ad_{ie}^T C_i Ad_{ie} w_e. \tag{A-23}$$

Further,

$$C_e = \sum_{i=0}^{m} Ad_{ie}^T C_i Ad_{ie}. \tag{A-24}$$

As shown in Fig. 15, we can obtain the external wrench $w_e$ and each wrench $w_i(i = 1, \dots, m)$ from the contact finger as

$$w_e = \sum_{i=1}^{m} w_i. \tag{A-25}$$

Substituting Eq. (A-18) into Eq. (A-25), we can obtain

$$w_e = \sum_{i=1}^{m} Ad_{ie}^{-1} w_i'. \tag{A-26}$$

By introducing the coordinate transformation matrix $Ad_{ie}$, we have

$$\sum_{i=1}^{m} T_i' = \Delta Ad_{ie} \Delta T_e = Ad_{ie}^{-T} T_e. \tag{A-27}$$

Through the stiffness matrix as

$$w_i' = K_i T_i' \tag{A-28}$$

where $K_i$ is the stiffness matrix of the $i$-th flexible element. Similarly, we can define the global stiffness matrix $K_e$ of the whole grasp system as

$$w_e = K_e T_e. \tag{A-29}$$

Substituting Eq. (A-28) and Eq. (A-29) into Eq. (A-26), we have

$$K_e T_e = \sum_{i=1}^{m} Ad_{ie}^{-1} K_i T_i' \tag{A-30}$$

which can be further simplified by substituting into Eq. (A-27) to obtain

$$K_e = \sum_{i=1}^{m} Ad_{ie}^{-1} K_i Ad_{ie}^{-T}. \tag{A-31}$$

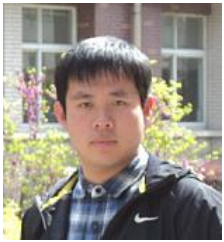

**Huixu Dong** received the B.Sc degree in mechatronics engineering from Harbin Institute of Technology in China, in 2013. He pursued Ph.D. at Robotics Research Centre of Nanyang Technological University, Singapore. Currently, he is a researcher in Robotics Institute of Carnegie Mellon University. His current research interests include robotic perception for grasping, optimization of grasping stability, robotics-oriented computer vision, and the navigation of mobile robot.

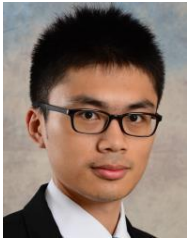

**Chen Qiu** received the PhD in Robotics from King's College London (KCL) in 2016. Previously, he received his B.S. degree in Spacecraft Design and Engineering from Beihang University, China in 2011. He is a postdoctoral fellow at the Robotics Research Centre, Nanyang Technological University. His research interests include theoretical kinematics and dynamics of robotics, robotics application in human-robot interaction.

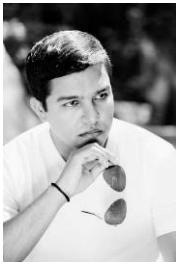

Dilip K. Prasad received the B.Tech. degree in computer science and engineering from the Indian Institute of Technology, Dhanbad, India, in 2003, and the Ph.D. degree from Nanyang Technological University, Singapore, in 2013, both in computer science and engineering. He is currently a Senior Research Fellow with Nanyang Technological University. He has authored more than 60 internationally peer-reviewed research articles. His current research interests include robotics perception and grasp, image processing, pattern recognition, and computer vision.

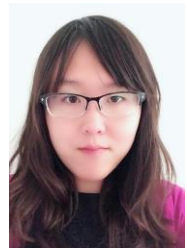

**Ye Pan** received her B.Sc degree in Communication and Information Engineering from Purdue/UESTC in 2010 and PhD degree in Computer Graphics from the University College London (UCL) in 2015. Currently, she is a postdoctoral associate at Disney Research Los Angeles. Her research interests include: mathematical modelling for robotics, robotics-human interaction, computer graphics.

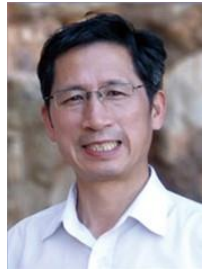

**Jiansheng Dai** received the B.Sc Degree, M.S. in mechanical engineering from Shanghai Jiaotong University in China in 1982 and 1984, respectively. He obtained his Ph.D degree in mechanical engineering from University of Salford, in England in 1993. he is a fellow of ASME, fellow of IEEE, fellow of IMechE and Chair of Mechanisms and Robotics at Centre for Robotics Research, King's College London, University of London. He has been working in the field of mechanisms and robotics in the past 26 years and published over 450 peer-reviewed papers including 230 journal papers and 4 books. Jian was a recipient of the ASME outstanding contribution award as the Conference Chair of the 36th ASME mechanisms and robotics conference held in Chicago. Jian is currently a Subject Editor of Mechanism and Machine Theory and an Associate Editor of IEEE Transactions on Robotics. His research interest is in reconfigurable mechanisms, dexterous mechanisms, end-effectors and multi-fingered hands.

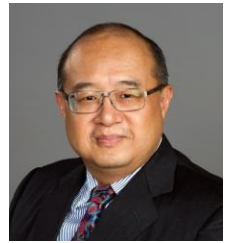

**I-Ming Chen** received the B.S. degree from National Taiwan University in 1986, and M.S. and Ph.D. degrees from California Institute of Technology, Pasadena, CA in 1989 and 1994 respectively. He is a full professor of the School of Mechanical and Aerospace Engineering, directions of Robotics Research Centre and Intelligent System Centre in Nanyang Technological University, Singapore. He is Fellow of ASME and Fellow of IEEE, General Chairman of 2017 IEEE International Conference on Robotics and Automation (ICRA2017). He is a senior editor of IEEE transaction on robotics. He also acts as the Deputy Program Manager of A*STAR SERC Industrial Robotics Program to coordinate project and activities under this multi-institutional program involving NTU, NUS, SIMTech, A*STAR I2R and SUTD. He is a member of the Robotics Task Force 2014 under the National Research Foundation. He works on many different topics in robotics, such as mechanism, actuator, human-robot interaction, perception and grasp, and industrial automation.